\documentclass[sigconf]{acmart}

\AtBeginDocument{%
  }

\usepackage{placeins}

\acmYear{2026}
\acmConference[Preprint]{Preprint}{2026}{}

\setcopyright{none}
\renewcommand\footnotetextcopyrightpermission[1]{}
\acmDOI{}
\acmISBN{}

\begin{document}

\title{Budget-Xfer: Budget-Constrained Source Language Selection for Cross-Lingual Transfer to African Languages}

\author{Tewodros Kederalah Idris}
\affiliation{
  \institution{Carnegie Mellon University Africa}
  \country{Rwanda}
}
\email{tidris@andrew.cmu.edu}

\author{Roald Eiselen}
\affiliation{
  \institution{North-West University}
  \country{South Africa}
}
\email{Roald.Eiselen@nwu.ac.za}

\author{Prasenjit Mitra}
\affiliation{
  \institution{Carnegie Mellon University Africa}
  \country{Rwanda}
}
\email{prasenjm@andrew.cmu.edu}

\renewcommand{\shortauthors}{Idris et al.}

\begin{abstract}
Cross-lingual transfer learning enables NLP for low-resource languages by leveraging labeled data from higher-resource sources, yet existing comparisons of source language selection strategies do not control for total training data, confounding language selection effects with data quantity effects. We introduce \textsc{Budget-Xfer}, a framework that formulates multi-source cross-lingual transfer as a budget-constrained resource allocation problem. Given a fixed annotation budget~$B$, our framework jointly optimizes which source languages to include and how much data to allocate from each. We evaluate four allocation strategies across named entity recognition and sentiment analysis for three African target languages (Hausa, Yoruba, Swahili) using two multilingual models, conducting 288 experiments. Our results show that (1)~multi-source transfer significantly outperforms single-source transfer (Cohen's $d = 0.80$--$1.98$), driven by a structural budget underutilization bottleneck; (2)~among multi-source strategies, differences are modest and non-significant; and (3)~the value of embedding similarity as a selection proxy is task-dependent, with random selection outperforming similarity-based selection for NER but not sentiment analysis.
\end{abstract}

\ccsdesc[500]{Computing methodologies~Information extraction}
\ccsdesc[300]{Computing methodologies~Language resources}
\ccsdesc[300]{Computing methodologies~Machine learning approaches}

\keywords{cross-lingual transfer, resource allocation, low-resource NLP, African languages, source language selection}

\maketitle

\section{Introduction}
\label{sec:introduction}

Cross-lingual transfer learning enables NLP systems for low-resource languages by training on labeled data from higher-resource source languages. This is particularly important for African languages, which remain severely underrepresented in NLP despite serving hundreds of millions of speakers~\cite{joshi2020state, adelani2021masakhaner}. A fundamental question is \emph{source language selection}: which source languages should be used for training, and how should data be distributed among them?

Several selection strategies have been explored, including typological distance from linguistic databases~\cite{littell2017uriel, dryer2013wals}, embedding similarity from multilingual models~\cite{lin2019choosing, eronen2023zero}, and multi-source transfer combining data from multiple sources~\cite{chen2019multi, lim2024analysis}. However, existing comparisons share a critical limitation: they do not control for total training data. Because source languages have varying dataset sizes, different strategies yield different total training examples, confounding language selection with data quantity.

Lim et al.~\cite{lim2024analysis} addressed this by equalizing total training size, but used only uniform allocation and focused on understanding \emph{why} multi-source transfer works rather than \emph{how to optimally distribute} a fixed budget. The question practitioners face is concrete: ``Given $N$ labeled sentences, which languages should I annotate and how much from each?''

We introduce \textsc{Budget-Xfer}, a framework that formulates multi-source cross-lingual transfer as a budget-constrained resource allocation problem. Given a fixed budget~$B$, we jointly optimize which source languages to include and how much data to allocate from each. By holding the total budget constant across all strategies, we isolate the effect of language selection and allocation decisions from data quantity. We compare four allocation strategies on NER and sentiment analysis across three African target languages using two multilingual models. Our contributions are:
\begin{enumerate}
    \item A budget-constrained formulation enabling fair comparison of allocation strategies under controlled data conditions. While prior work has controlled total data size~\cite{lim2024analysis}, we are the first to systematically compare allocation strategies, treating the distribution of data across sources as the variable of interest.
    \item Four allocation strategies including a diversity-aware approach that balances source-target similarity with inter-source redundancy.
    \item Budget-controlled experiments (288 runs) yielding actionable findings: multi-source transfer strongly outperforms single-source due to a structural budget underutilization bottleneck, differences among multi-source strategies are modest, and the value of similarity-based selection is task-dependent.
\end{enumerate}

\section{Related Work}
\label{sec:related}

Source language selection for cross-lingual transfer has been studied through multiple lenses. Lin et al.~\cite{lin2019choosing} introduced LangRank, framing source selection as a ranking problem. Eronen et al.~\cite{eronen2023zero} showed that linguistic similarity metrics correlate with transfer performance and that optimal source selection yields significant gains over defaulting to English. These methods address \emph{which} source to select but do not consider how to distribute a fixed budget across multiple sources.

Multi-source transfer has been studied by Chen et al.~\cite{chen2019multi}, who proposed an adversarial mixture-of-experts framework. Most closely related to our work, Lim et al.~\cite{lim2024analysis} controlled total training size across conditions and found that multi-source training increases embedding space mingling. However, they used only uniform allocation and focused on understanding the mechanism rather than optimizing the allocation itself. We build on their insight that data quantity must be controlled, and go further by treating the allocation vector as the variable to optimize.

Data allocation across languages has been studied for multilingual \emph{pre-training}~\cite{conneau2020unsupervised}, but these approaches operate over unlabeled data and optimize for general-purpose representations. Our work addresses allocation for supervised \emph{fine-tuning}, where the budget is orders of magnitude smaller, each example is costly to annotate, and the allocation directly determines downstream task performance.

\section{Methodology}
\label{sec:methodology}

\subsection{Problem Formulation}

Given a target language $t$, a pool of $n$ candidate source languages $\mathcal{S} = \{s_1, \ldots, s_n\}$, and a fixed budget $B$ (in training sentences), we seek an allocation vector $\mathbf{a} = [a_1, \ldots, a_n]$ maximizing transfer performance:
\begin{equation}
\label{eq:objective}
\mathbf{a}^* = \arg\max_{\mathbf{a}} \; P(t \mid \mathbf{a})
\quad \text{s.t.} \quad \sum_{i=1}^{n} a_i = B, \;\; a_i \geq 0
\end{equation}
Since exhaustive search is infeasible, we use a pre-computed utility proxy to approximate source value. Following prior work showing \textit{cosine\_gap} predicts cross-lingual transfer ($\rho = 0.4$--$0.6$)~\cite{idris2026embedding}, we compute pairwise similarities using 1,012 parallel sentences per language from FLORES-200~\cite{nllb2022}:
\begin{equation}
\text{cosine\_gap}(s, t) = \overline{\cos(\mathbf{h}_s, \mathbf{h}_t)}_{\text{aligned}} - \overline{\cos(\mathbf{h}_s, \mathbf{h}_t)}_{\text{misaligned}}
\end{equation}

\subsection{Allocation Strategies}
\label{sec:strategies}

We compare four strategies (Table~\ref{tab:strategies}) varying in source selection and budget distribution. When allocations exceed a source's available data, a cap-and-redistribute procedure ensures total training data is exactly $B$ for all strategies.

\begin{table}[t]
\centering
\caption{Allocation strategies. $K{=}5$ for all multi-source strategies.}
\label{tab:strategies}
\small
\begin{tabular}{lll}
\toprule
\textbf{Strategy} & \textbf{Selection} & \textbf{Allocation} \\
\midrule
\textsc{All-from-Best} & Top 1 by sim & Entire budget \\
\textsc{Top-K-Prop.} & Top $K$ by sim & Proportional to sim \\
\textsc{Random-K} & $K$ random & Equal split \\
\textsc{Diversity-Aware} & Greedy diverse $K$ & Proportional to sim \\
\bottomrule
\end{tabular}
\end{table}

\textsc{All-from-Best} allocates the entire budget to the single most similar source ($a_{s^*} = B$, $s^* = \arg\max_{s} \text{sim}(s,t)$), serving as the single-source baseline. This tests whether concentrating all resources on the most promising source outperforms distributing data across multiple languages.

\textsc{Top-K-Proportional} selects the $K$ most similar sources and allocates proportionally to their similarity scores:
\begin{equation}
a_i = B \cdot \frac{\text{sim}(s_i, t)}{\sum_{s_j \in \text{Top-}K} \text{sim}(s_j, t)}
\end{equation}
More similar sources receive proportionally more training data, testing whether weighting by similarity improves over uniform distribution.

\textsc{Random-K} selects $K$ sources uniformly at random with equal allocation ($a_i = B/K$). This uninformed baseline uses no similarity information; if similarity-informed strategies consistently outperform \textsc{Random-K}, it validates the utility proxy under budget constraints.

\textsc{Diversity-Aware} modifies source selection to penalize redundancy among chosen sources. Sources are selected iteratively by maximizing:
\begin{equation}
\text{score}(s) = \text{sim}(s, t) - \alpha \cdot \max_{s' \in \mathcal{C}} \text{sim}(s, s')
\end{equation}
where $\mathcal{C}$ is the set of already-selected sources and $\alpha = 0.5$ controls the diversity penalty. When $\alpha = 0$, this reduces to standard top-$K$ selection. In the first round, $\mathcal{C} = \emptyset$ and $\text{score}(s) = \text{sim}(s, t)$, so the most similar source is always selected first. After selection, budget is allocated proportionally to raw similarity, separating the roles of diversity (affecting selection) and similarity (determining allocation). Comparing \textsc{Diversity-Aware} to \textsc{Top-K-Proportional} isolates the effect of diversity-based selection, since both use the same allocation rule but may select different sources.

We also evaluated Top-K-Uniform and Greedy-Marginal strategies; results were consistent with the four strategies reported here.

\section{Experimental Setup}
\label{sec:experiments}

\paragraph{Tasks.} We evaluate on two standard NLP tasks for African languages. For NER, we use MasakhaNER 2.0~\cite{adelani2022masakhaner}, which provides entity-annotated data in 20 African languages with standard categories (PER, ORG, LOC, DATE); the evaluation metric is entity-level F1 (micro-averaged) computed with \texttt{seqeval}~\cite{seqeval}, requiring exact span and type match. For sentiment analysis, we use AfriSenti~\cite{muhammad2023afrisenti}, covering 14 African languages with three classes (positive, negative, neutral); the evaluation metric is weighted F1 to account for class imbalance.

\paragraph{Languages.} We select three typologically diverse target languages representing different language families: Hausa (Afro-Asiatic, Chadic branch), Yoruba (Niger-Congo, Volta-Niger branch), and Swahili (Niger-Congo, Bantu branch). These differ in family affiliation, morphological typology, and geographic distribution. For each target, the candidate source pool consists of all other languages in the respective dataset: 19 candidates for NER and 13 for sentiment.\footnote{One AfriSenti language (Nigerian Pidgin) lacks FLORES-200 parallel data for similarity computation; it receives a similarity score of zero and is effectively excluded from similarity-informed strategies but remains eligible for random selection.}

\paragraph{Models.} To assess robustness across architectures, we evaluate with two multilingual models: Serengeti~\cite{adebara2023serengeti}, covering 517 African languages, and AfroXLMR~\cite{alabi2022adapting}, an XLM-RoBERTa adaptation for African languages. Both are used for computing embedding similarities and for fine-tuning.

\paragraph{Budgets.} We evaluate at two budget levels per task, calibrated to dataset characteristics: NER uses 10K (low) and 15K (high) sentences, while sentiment uses 5K (low) and 10K (high), reflecting the smaller per-language datasets in AfriSenti.

\paragraph{Training.} All models are fine-tuned with AdamW~\cite{loshchilov2019decoupled} ($\text{lr}=2{\times}10^{-5}$, weight decay 0.01), batch size 16, max sequence length 128, and up to 10 epochs with early stopping (patience 3). Data is sampled without replacement according to each strategy's allocation vector, concatenated and shuffled without language markers, with 10\% held out for validation. Evaluation is zero-shot: models train on source data only and are tested on the target language. Each of 2 tasks $\times$ 3 targets $\times$ 2 budgets $\times$ 2 models $\times$ 4 strategies = 96 configurations is run with 3 seeds (42, 43, 44), totaling 288 main runs plus 144 with additional strategies (432 total). We report paired $t$-tests with Bonferroni correction and Cohen's $d$ effect sizes.

\section{Results}
\label{sec:results}

Tables~\ref{tab:ner_results} and~\ref{tab:sentiment_results} present results on Serengeti (primary model). We organize analysis around three findings.

\begin{table}[t]
\centering
\caption{NER results (F1, Serengeti). Best per column in bold.}
\label{tab:ner_results}
\footnotesize
\setlength{\tabcolsep}{3pt}
\begin{tabular}{lcccccccc}
\toprule
& \multicolumn{4}{c}{Low (10K)} & \multicolumn{4}{c}{High (15K)} \\
\cmidrule(lr){2-5} \cmidrule(lr){6-9}
Strategy & Hau & Yor & Swa & Avg & Hau & Yor & Swa & Avg \\
\midrule
All-Best & .643 & .526 & .698 & .623 & .635 & .504 & .688 & .609 \\
T5-Prop & .689 & \textbf{.550} & .743 & .660 & .701 & \textbf{.566} & .733 & .666 \\
Rand-5 & .706 & .546 & \textbf{.801} & \textbf{.684} & \textbf{.713} & .564 & \textbf{.783} & \textbf{.686} \\
Div-Aw & \textbf{.716} & .490 & .790 & .665 & .711 & .493 & .779 & .661 \\
\bottomrule
\end{tabular}
\vspace{0.5mm}

\raggedright\scriptsize{All-Best = All-from-Best, T5-Prop = Top-5-Proportional, Div-Aw = Diversity-Aware.}
\end{table}

\begin{table}[t]
\centering
\caption{Sentiment results (weighted F1, Serengeti). Best per column in bold.}
\label{tab:sentiment_results}
\footnotesize
\setlength{\tabcolsep}{3pt}
\begin{tabular}{lcccccccc}
\toprule
& \multicolumn{4}{c}{Low (5K)} & \multicolumn{4}{c}{High (10K)} \\
\cmidrule(lr){2-5} \cmidrule(lr){6-9}
Strategy & Hau & Yor & Swa & Avg & Hau & Yor & Swa & Avg \\
\midrule
All-Best & .199 & .196 & \textbf{.553} & .316 & .262 & .208 & \textbf{.553} & .341 \\
T5-Prop & .570 & \textbf{.466} & .469 & .502 & .529 & \textbf{.418} & .485 & .478 \\
Rand-5 & .542 & .347 & .373 & .421 & .548 & .355 & .420 & .441 \\
Div-Aw & \textbf{.574} & .428 & .505 & \textbf{.502} & \textbf{.550} & .414 & .518 & \textbf{.494} \\
\bottomrule
\end{tabular}
\vspace{0.5mm}

\raggedright\scriptsize{Abbreviations as in Table~\ref{tab:ner_results}.}
\end{table}

\subsection{Multi-Source vs.\ Single-Source Transfer}
\label{sec:rq1}

The most consistent finding is that multi-source strategies substantially outperform single-source transfer across all conditions. \textsc{All-from-Best} ranks last in mean F1 for both tasks and both budget levels, with an overall mean of 0.472 compared to 0.572 for the average multi-source strategy. Paired comparisons between \textsc{All-from-Best} and \textsc{Top-5-Proportional} (Table~\ref{tab:comparisons}) yield large effect sizes: Cohen's $d = 1.43$ and $1.98$ for NER ($p < 0.01$), and $d = 0.86$ and $0.80$ for sentiment ($p < 0.05$).

The key driver of this gap is \emph{budget underutilization}. Because \textsc{All-from-Best} concentrates the entire budget on a single source, it is constrained by that language's available data. Across all runs, \textsc{All-from-Best} utilized only 57\% of the allocated budget on average (range: 27\%--100\%), while all multi-source strategies consistently utilized the full budget. The effect is most severe for sentiment analysis targeting Hausa, where the best source (Swahili) has only 1,810 available sentences, yielding 36\% utilization at the 5K budget and just 18\% at the 10K budget. Multi-source strategies distribute the budget across sources, each contributing within its available capacity, and the cap-and-redistribute procedure ensures full budget utilization.

The one exception confirms this explanation: for Swahili sentiment, the best source (Hausa) provides sufficient data for 100\% budget utilization, and \textsc{All-from-Best} achieves the highest F1 (0.553). This indicates that the single-source disadvantage is driven primarily by budget underutilization rather than an inherent limitation of concentrated training.

\begin{table}[t]
\centering
\caption{Key pairwise comparisons (paired $t$-tests, Serengeti). Positive $\Delta$ favors the second strategy.}
\label{tab:comparisons}
\footnotesize
\setlength{\tabcolsep}{2.5pt}
\begin{tabular}{llccccc}
\toprule
Comparison & Task-Bud. & $n$ & $\Delta$ & 95\% CI & $p$ & $d$ \\
\midrule
Single vs Multi & NER-L & 9 & +.038 & [+.018,+.058] & .003** & +1.43 \\
(All-Best vs & NER-H & 9 & +.057 & [+.035,+.079] & .000*** & +1.98 \\
T5-Prop) & Snt-L & 9 & +.185 & [+.019,+.351] & .033* & +0.86 \\
 & Snt-H & 9 & +.137 & [+.005,+.269] & .044* & +0.80 \\
\midrule
Std vs Diverse & NER-L & 9 & +.005 & [$-$.035,+.045] & .777 & +0.10 \\
(T5-Prop vs & NER-H & 9 & $-$.006 & [$-$.052,+.041] & .793 & $-$0.09 \\
Div-Aw) & Snt-L & 9 & +.000 & [$-$.042,+.043] & .984 & +0.01 \\
 & Snt-H & 9 & +.017 & [$-$.049,+.082] & .572 & +0.20 \\
\midrule
Informed vs Rand & NER-L & 9 & +.024 & [$-$.006,+.054] & .155 & +0.52 \\
(T5-Prop vs & NER-H & 9 & +.020 & [$-$.008,+.048] & .205 & +0.46 \\
Rand-5) & Snt-L & 9 & $-$.081 & [$-$.181,+.019] & .152 & $-$0.53 \\
 & Snt-H & 9 & $-$.037 & [$-$.107,+.034] & .337 & $-$0.34 \\
\bottomrule
\end{tabular}
\vspace{0.5mm}

\raggedright\scriptsize{$^{*}p<.05$, $^{**}p<.01$, $^{***}p<.001$. L=Low budget, H=High budget, Snt=Sentiment.}
\end{table}

\subsection{Allocation Strategy Comparison}
\label{sec:rq2}

Given that multi-source transfer is clearly preferable, we next examine whether the specific allocation rule and selection criterion matter. Among multi-source strategies, overall mean F1 scores on Serengeti are tightly clustered: \textsc{Diversity-Aware} (0.581), \textsc{Top-5-Proportional} (0.577), and \textsc{Random-5} (0.558). The controlled comparison between \textsc{Top-5-Proportional} and \textsc{Diversity-Aware} (Table~\ref{tab:comparisons}), which isolates the effect of diversity-based selection while holding the allocation rule constant, shows no significant difference in any condition ($|d| \leq 0.20$, all $p > 0.57$).

We also tested Top-K-Uniform and Greedy-Marginal allocation in our full experimental design. Top-K-Uniform showed no significant difference from Top-K-Proportional ($|d| \leq 0.52$, all $p > 0.15$), indicating that the allocation rule (uniform vs.\ proportional) does not matter when the same sources are selected. Greedy-Marginal, which performs selection and allocation jointly through iterative batch allocation with diminishing returns, was marginally significant for NER at low budget ($d = 0.77$, $p = 0.049$) but non-significant in all other conditions.

Win rates across all seed-level conditions are distributed remarkably evenly among multi-source strategies: \textsc{Random-5}, \textsc{Top-5-Proportional}, and \textsc{Diversity-Aware} each win 28\% of conditions, with \textsc{All-from-Best} winning 17\%. This suggests that once the decision to use multiple sources is made, the exact allocation mechanism has a smaller effect than the multi-source decision itself.

\subsection{Task-Dependent Value of Similarity}
\label{sec:rq3}

Although multi-source strategies perform similarly on aggregate, a task-dependent pattern emerges when comparing informed and uninformed selection.

\paragraph{NER.} \textsc{Random-5} achieves the highest mean F1 at both budget levels (0.684 at 10K, 0.686 at 15K), outperforming all similarity-informed strategies. The controlled comparison between \textsc{Top-5-Proportional} and \textsc{Random-5} (Table~\ref{tab:comparisons}) shows random selection is directionally better at both budget levels (low: $\Delta = +0.024$, $d = 0.52$, $p = 0.16$; high: $\Delta = +0.020$, $d = 0.46$, $p = 0.21$), though neither difference reaches significance.

Examining selected sources reveals why: the cosine\_gap metric assigns high scores to typologically related languages, causing top-5 selection to concentrate within Bantu languages. For Hausa as target, \textsc{Top-5-Proportional} selects Swahili, Kinyarwanda, Igbo, Zulu, and Chichewa, of which four are Bantu. In contrast, \textsc{Random-5} (seed 42) selects Fon, Bambara, Mossi, Luo, and Zulu, spanning five distinct branches: Gbe, Mande, Gur, Nilotic, and Bantu. For NER, where the model must learn to recognize diverse entity boundary patterns and entity types, exposure to varied linguistic structures appears more beneficial than concentrated exposure to similar structures.

\paragraph{Sentiment.} The pattern reverses. \textsc{Top-5-Proportional} and \textsc{Diversity-Aware} consistently outperform \textsc{Random-5} by 4--8 F1 points at both budget levels (Table~\ref{tab:comparisons}: low $d = -0.53$, high $d = -0.34$), and informed strategies rank above \textsc{Random-5} in all conditions. Although these differences do not reach significance with our sample size, the consistent directionality suggests that linguistically similar sources provide more transferable signal for sentiment analysis.

\paragraph{Task Specificity.} These findings caution against treating embedding similarity as a universal selection criterion. For token-level tasks like NER, where generalization requires recognizing diverse surface patterns, similarity-based selection may be counterproductive by over-concentrating on redundant sources. For sentence-level semantic tasks like sentiment analysis, where the relevant linguistic properties are more closely aligned with what embedding similarity captures, informed selection provides a consistent directional advantage. This suggests practitioners should consider the target task when choosing a selection strategy.

\subsection{Budget Effects and Cross-Model Robustness}

Increasing the budget from low to high yields minimal performance changes: $-0.008$ mean F1 for NER (10K to 15K) and $+0.012$ for sentiment (5K to 10K). For NER, performance is flat or slightly decreasing, suggesting that 10,000 training sentences already reach a saturation point for cross-lingual transfer with these models. For sentiment, the modest gain indicates that 5,000 sentences capture most of the transferable signal. These results suggest that practitioners can achieve near-optimal transfer performance with relatively modest annotation budgets, provided data is distributed across multiple sources rather than concentrated in a single source.

To assess generalizability across model architectures, we repeat all experiments with AfroXLMR (Table~\ref{tab:cross_model}). While the top-ranked strategy differs between models in some conditions, the overall patterns are consistent. For NER, \textsc{Random-5} leads on Serengeti while \textsc{Top-5-Proportional} leads on AfroXLMR, but multi-source strategies dominate on both. For sentiment, \textsc{Diversity-Aware} ranks first on both models in 3 of 4 conditions. \textsc{All-from-Best} consistently ranks last or second-to-last on both models. The core finding that multi-source transfer outperforms single-source transfer holds across architectures, as does the observation that differences among multi-source strategies are small relative to the single-vs-multi gap.

\begin{table}[t]
\centering
\caption{Cross-model strategy rankings (top 3). The competitive set is consistent despite different top-1 picks.}
\label{tab:cross_model}
\footnotesize
\setlength{\tabcolsep}{3pt}
\begin{tabular}{llll}
\toprule
Task & Bud. & Serengeti & AfroXLMR \\
\midrule
NER & Low & \textbf{Rand.} $>$ Div. $>$ T5-P & \textbf{T5-P} $>$ Rand. $>$ Div. \\
NER & High & \textbf{Rand.} $>$ T5-P $>$ Div. & \textbf{T5-P} $>$ Rand. $>$ Div. \\
Sent. & Low & \textbf{Div.} $>$ T5-P $>$ Rand. & \textbf{Div.} $>$ T5-P $>$ A-Best \\
Sent. & High & \textbf{Div.} $>$ T5-P $>$ Rand. & \textbf{Div.} $>$ T5-P $>$ Rand. \\
\bottomrule
\end{tabular}
\vspace{0.5mm}

\raggedright\scriptsize{T5-P = Top-5-Prop., Div. = Diversity-Aware, Rand.\ = Random-5, A-Best = All-from-Best. Best in bold. All-from-Best ranks last or second-to-last in all conditions on both models.}
\end{table}

\section{Conclusion}
\label{sec:conclusion}

We introduced \textsc{Budget-Xfer}, a framework for budget-constrained source language selection in cross-lingual transfer, addressing a known methodological limitation in prior work where different strategies yield different total training data.

Our experiments across 288 training runs yield three actionable findings. First, practitioners should always distribute data across multiple source languages rather than concentrating on a single source; single-source strategies face a structural budget underutilization bottleneck where, on average, only 57\% of the allocated budget is used. Second, once data is distributed across multiple sources, the specific allocation mechanism has a comparatively small effect, with no single multi-source strategy consistently dominating. Third, the value of embedding similarity as a source selection proxy is task-dependent: for NER, similarity-based selection concentrates sources within typologically related clusters, reducing the linguistic diversity that benefits token-level generalization, while for sentiment analysis, linguistically similar sources provide more transferable signal.

Several limitations suggest directions for future work. Our evaluation covers two tasks and three target languages; extending to additional tasks (e.g., dependency parsing, question answering) and a broader set of target languages would strengthen generalizability. We used a fixed $K=5$ for multi-source strategies, but preliminary results with a greedy marginal approach that distributed data across 14--17 sources suggest that optimizing $K$ jointly with the allocation strategy may yield further gains. Finally, our budget formulation treats all training sentences as equally costly to annotate; incorporating variable annotation costs across languages and tasks would better reflect real-world constraints.


\clearpage
\bibliographystyle{ACM-Reference-Format}
\bibliography{sample-base}

\end{document}